\documentclass[preprint,12pt]{elsarticle}

\usepackage{graphicx}
\usepackage{amssymb}
\usepackage{commath}

\usepackage{lineno}

\usepackage{amsthm}
\theoremstyle{plain}
\newtheorem{thm}{Theorem}[section] 

\theoremstyle{definition}
\newtheorem{defn}[thm]{Definition} 
\newtheorem{exmp}[thm]{Example} 

\journal{Transportation Research Part C: Emerging Technologies }

\begin{document}

\begin{frontmatter}


\title{Real-time Travel Time Estimation Using Matrix Factorization\tnoteref{0}}
\address{Tehran, Iran}



\author[1]{Ebrahim Badrestani\fnref{10}}
\address[1]{Sharif University of Technology}
\ead{badrestani@ce.sharif.edu}

\author[2]{Behnam Bahrak\fnref{10}}
\address[2]{University of Tehran}
\ead{bahrak@ut.ac.ir}

\author[2]{Ali Elahi\fnref{10}}
\ead{ali.elahi@ut.ac.ir}

\author[2]{Adib Faramarzi\fnref{10}}
\ead{adib.faramarzi@ut.ac.ir}

\author[2]{Pouria Golshanrad\fnref{10}}
\ead{pouria.golshanrad@ut.ac.ir}

\author[3]{Amin Karimi Monsefi\corref{cor2}\fnref{10}}
\address[3]{Shahid Beheshti University}
\ead{A_karimimonsefi@sbu.ac.ir}

\author[2]{Hamid Mahini\fnref{10}}
\ead{hamid.mahini@ut.ac.ir}

\author[2]{Armin Zirak\corref{cor2}\fnref{10}}
\ead{armin.zirak97@ut.ac.ir}

\fntext[10]{All names are listed alphabetically}

\cortext[cor2]{Corresponding authors}

\tnotetext[0]{This research is fully supported by Tap30 co, an online taxi-hailing company in Iran}

\begin{abstract}
Estimating the travel time of any route is of great importance for trip planners, traffic operators, online taxi dispatching and ride-sharing platforms, and navigation provider systems. With the advance of technology, many traveling cars, including online taxi dispatch systems' vehicles are equipped with Global Positioning System (GPS) devices that can report the location of the vehicle every few seconds. This paper uses GPS data and the Matrix Factorization techniques to estimate the travel times on all road segments and time intervals simultaneously. We aggregate GPS data into a matrix, where each cell of the original matrix contains the average vehicle speed for a segment and a specific time interval. One of the problems with this matrix is its high sparsity. We use Alternating Least Squares (ALS) method along with a regularization term to factorize the matrix. Since this approach can solve the sparsity problem that arises from the absence of cars in many road segments in a specific time interval, matrix factorization is suitable for estimating the travel time. Our comprehensive evaluation results using real data provided by one of the largest online taxi dispatching systems in Iran, shows the strength of our proposed method. 

\end{abstract}


\end{frontmatter}

\section{Introduction}
\label{S:1}

Travel time estimation in a city is a challenging and important task used for trip planners, traffic operators, online taxi dispatch and ride-sharing platforms, and navigation provider systems. The 2007 urban mobility report \cite{hunter2009path} states that traffic congestion causes 4.2 billion hours of extra travel in the United States every year, which accounts for 2.9 billion extra gallons of fuel, which cost taxpayers an additional \$78 billion. On the other hand, travel time estimation can help traffic operators have a deeper understanding of the current traffic flow and online taxi dispatch systems to estimate ride fares more accurately based on the current traffic conditions of a city. Travel time information, as a by-product, can also be used in navigational systems to suggest the fastest route with the least amount of traffic. 

Travel time data are collected through different means such as inductive loops, surveillance cameras, or mobile devices and vehicles equipped with a Global Positioning System (GPS) device. Since loop detectors and surveillance cameras are not available in all regions of a road network, especially the regions that are less congested, many recent proposed methods focus on using GPS data to estimate or predict the travel times \cite{hunter2009path,lin1999experimental,li2002link,amin2008mobile}. These devices can be used to communicate the current location of a moving vehicle along with time of the day to a central server every few seconds. The result is a massive data-set of time-based GPS trajectories.

Although there has been a focus on the problem of travel time estimation in the recent years \cite{wang2019simple,li2018multi,gentili2018review,zhu2018urban}, the task is still challenging, and many unresolved issues remain. Most of these papers focus on estimating travel times which are related to exceptional routes such as highways, and many of them consider unrealistic assumptions such as lack of data sparsity in the environment which is an issue that impacts the performance of the estimation method significantly. Another disadvantage is high computation complexity of most proposed methods, which in some cases can be impractical in real-world scenarios or for real-time applications \cite{olszewski2018assessing,hou2018estimating,oh2018short,lee2018evaluating}.

Motivated by these challenges, we study the travel time estimation in Tehran road network using the GPS data gathered from moving taxis of an online taxi dispatch system, in an unconditioned environment that suffers from data sparsity, and propose a method that is not restricted to highways or arterial roads and can be performed efficiently in a real-time manner.

We use matrix factorization alongside Alternating Least Squares (ALS) method to estimate the travel time of each road segments in a given time interval. The final matrix is computed in an iterative fashion with the historical data and then further improved using recent real-time data. Unlike many other methods that need to estimate the travel time on each segment individually, this method can propose estimations for all segments at all time-intervals, simultaneously \cite{paatero1994positive,koren2009matrix}.

The rest of this paper is organized as follows. In Section II, we discuss multiple travel time estimation techniques related to our work. We introduce matrix factorization with ALS method that is used to estimate travel times, in Section III. Section IV includes our evaluation results for the proposed method, and finally, Section V concludes the paper.

\section{Related Work}
\label{S:3}

Proposed methods for travel time estimation use different types of data for this purpose. Older models have used the Kalman Filter technique alongside the data provided by loop detectors and probe vehicles to estimate the travel times using all similar trips existing in historical data \cite{rice2004simple}.
Modern approaches for travel time estimation can be classified into four main categories:

\begin{itemize}
\item Origin-Destination-based approaches
\item Segment-based approaches
\item Sub-path-based approaches
\item Neural Network approaches
\end{itemize}

\subsection{Origin-Destination-based approaches}
Origin-Destination-Based approaches use the location and time of trips' origins and destinations, instead of using all information about road segments and the GPS points. Some methods first tessellate the city with horizontal and vertical guidelines and then estimate the travel times based on the tiles that the origin and destination of a trip fall into. Other similar statistical models have been proposed to estimate the travel times based on the trips that have similar origin/destination pair to the query trip \cite{huang2016optimal,yang2017origin,silva2015odcrep}.

\begin{figure}
\centering
\includegraphics[totalheight=6cm]{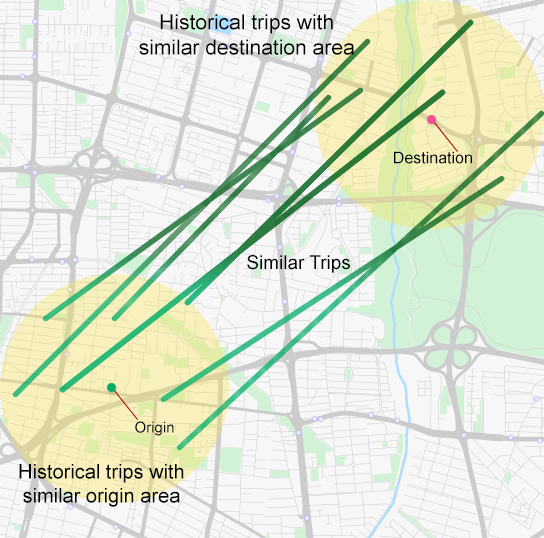}
\caption{Similar Trips Based on the Origin and Destination}
\label{fig:tehran_od}
\end{figure}

Hongijan et al. proposed a solution that averages all the travel times of similar trips based on the origin and destination locations after regularizing them \cite{wang2019simple}. This solution is analyzed thoroughly in our paper, and compared to our proposed method. In \cite{jindal2017unified}, Jindal et al. propose a multi-layer feed-forward neural network for travel time estimation called Spatio-Temporal Neural Network (ST-NN). It takes the latitudes and longitudes of the origin and destination points as input and combines them with the time information to estimate the trip duration. These methods have a hard time utilizing the raw features such as origin, destination, departure time, and need to learn feature representation to improve their models.

\subsection{Segment-based approaches}

These methods estimate the travel time on each individual segment of the path, assuming that the path of the trip is given or calculated beforehand, and then sum the estimated travel times of each segment to get the path travel time. Proposed methods use different data to train their models, including loop-detectors that gather the data of vehicles passing over them, surveillance cameras, and moving cars equipped with a GPS module from which each location is sent to a central server. 

Some methods try to infer vehicle speed from loop detectors and then estimate travel time on each individual path segment based on those recordings.

Nakata et al. \cite{nakata2004mining} treats the data of travel time as time series, and employ statistical models such as the autoregressive (AR) and state-space models to obtain precise estimation function.

\subsection{Sub-path-based approach}

These methods try to estimate the time of the whole path by the similar sub-paths of historical trips. 

\begin{figure}
\centering
\includegraphics[totalheight=6cm]{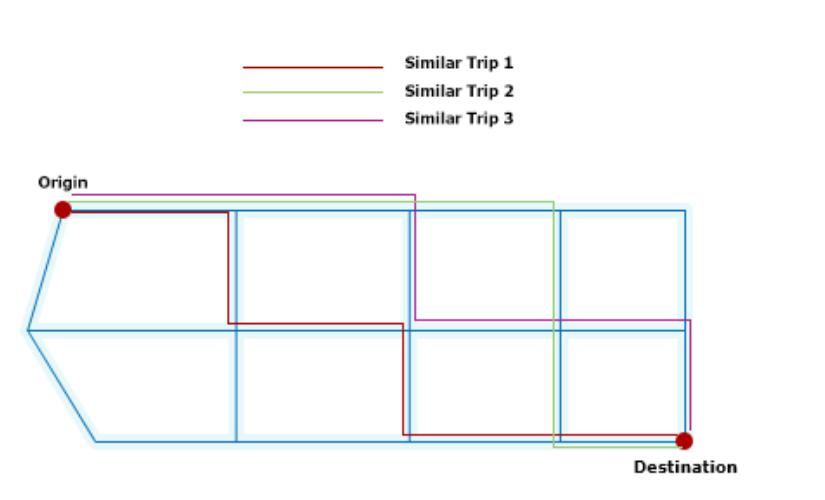}
\caption{Trips with Similar Sub-Paths}
\label{fig:multipaths}
\end{figure}

Rahmani et al. propose concatenating the sub-paths to estimate accurate travel times \cite{rahmani2013route}. Hongijan et al. mine the frequent patterns in the sub-paths and then find the optimal way to concatenate those sub-paths \cite{wang2014travel}. These methods have two major drawbacks for real-time usage: (1) finding the similar sub-paths can be computationally inefficient, especially when we want to estimate travel time in a real-time manner, and (2) there might not be enough sub-paths to take into account. This problem can have a deeper impact on accuracy when we have a lot of segments that are not used frequently and have infrequent or no data available.

\subsection{Neural Network approaches}

These approaches use different types of inputs and train a neural network model to estimate the travel time of a trip using historical data. Wu et al. model trajectory data with a recurrent neural network (RNN), capturing long-term historical data dependencies and patterns \cite{wu2017modeling}. Gao et al. try to represent the underlying semantics of mobility patterns by using RNN with embeddings \cite{gao2017identifying}. Zhang et al. partition the whole network into $N \times N$ disjoint but equal-sized grids, like what was done in Origin-Destination-Based approaches, and feed this data into a multi-layer neural network with short-term and long-term traffic feature extraction\cite{zhang2018deeptravel}. In general, neural network approaches achieve a better accuracy comparing to statistical methods, but they require a significant amount of data to thrive. Our data dimension is rather large compared to its number of records thus, using the neural networks model could not extract patterns and relations between the features efficiently.

\begin{figure}
\centering
\includegraphics[totalheight=7cm]{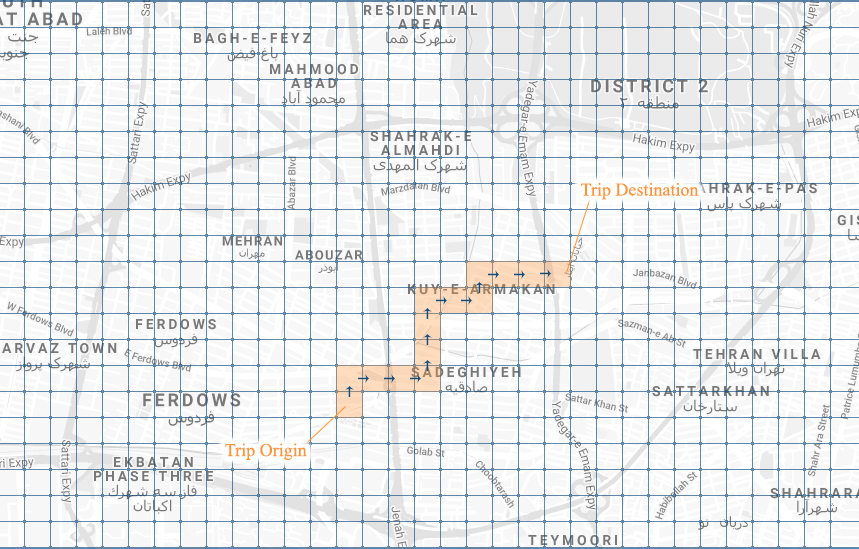}
\caption{Partitioning of Tehran}
\label{fig:terhan_grid}
\end{figure}

The model proposed in \cite{zheng2006short} feed the data from 15 minutes intervals to a neural network and attempts to predict the volume for the next 15 minutes. Although the accuracy of the proposed method is high, the experiments are limited, and the validation is only done on a three-point express-way during daytime hours.

Some methods explore physical relationships between travel time and other traffic factors such as traffic condition, weather, signal timings, and so on, to have a more accurate travel time estimation. However, these kinds of additional data might not always be available.

It should be noticed that some of the proposed methods do not perform well in real-time scenarios, for two reasons:
\begin{itemize}
\item Parameter decisions in neural networks, efficient pattern finding, and matching in sub-pattern based approaches and coefficient estimation in statistical models all need to be dynamic, due to the nature of the real-time applications.
\item The parameters for these approaches are trained for specific scenarios defined for their testbeds, e.g. highways and intersections, which leads to overfitting in an all-purpose scenario.
\end{itemize}

\section{Methodology}
\label{S:4}

We are going to introduce the architecture of a general matrix factorization model which is used for estimating the speed of segments in the graph of a base-map. Base-map is a directed graph which consists of nodes and segments. Each segment connects to two nodes with a direction. For each week, we build a matrix, where its rows represent road segments, and each column is responsible for a time interval of the week. Each cell of this matrix is filled with observations of the relevant segment and time interval of that week. Then, we use the average of these weekly matrices to build the aggregation matrix. Our goal is capturing the correlation among segments at different times. We utilize the matrix factorization (MF) method to find this correlation. MF helps us in solving the sparsity issue and reducing the noise of data. The MF model will be applied to the aggregation matrix to estimate the average vehicle speed in each segment. After that, we use real-time data to improve our estimation. We build a vector of the real-time speed in segments and apply the MF model on this vector to estimate the final speed for segments.

\subsection{Data Gathering}\label{sec:dg}
In this stage, we collect and match the locations of the drivers to the base-map and calculate the speed of the vehicles on each segment. The drivers move along the segments and send their locations coordination for each time interval. Since GPS data can have a lot of noise, especially in places with tall buildings and other obstacles, we map-match every reported GPS data to the nearest segment, as the data arrives.

We consider the locations of the vehicles batch by batch. In each batch, we have some locations sequences of the vehicles. After mapping the locations to their related segments, we calculate the average speed between each pair of the locations. Suppose we have two locations $l_1$ and $l_2$ which a driver sent at timestamps $t_1$ and $t_2$, respectively. Then we use the formula \ref{eq:dg:1} to calculate the speed of the vehicle between these locations. In this formula, $||l_2-l_1||$ means the haversine distance between $l_1$ and $l_2$.
\begin{equation}
    \textbf{speed} = \frac{||l_2-l_1||}{t_2-t_1}
    \label{eq:dg:1}
\end{equation}

\subsection{Data Aggregation}\label{sec:da}
As declared in the previous subsection, the time-segment matrix is generated with real-time data using the reported data from GPS-equipped devices passing through the segments. Many data cells do not have any data since not all segments contain a moving GPS device (e.g., a vehicle) in all-time intervals. If multiple cars have passed a certain segment in a time interval, their average speed would be used, and if there were no cars that have passed a segment in a time interval, the speed would be considered zero. Now, we have a matrix which the $\textbf{cell}_{ij}$ represents our estimated speed of segment $i$ at time interval $j$. 

One of the main issues when working with travel time estimation is the negative effect of data sparsity on accuracy. If we use small time-intervals, e.g., 5 seconds, there are a lot of cells containing zero, and if we use large time intervals, e.g. 12 hours, many zero cells would be eliminated, but the result is not reliable because of the nature of traffic and how it changes in practice.

To fix the data sparsity issue and also denoising the values of a matrix, we start with a low-rank approximation of the time-segment matrix T. The average speed of vehicles in a segment and at a specific time interval can be modeled through the inner product of the time feature vector and the segment feature vector.

\subsection{Problem Definition}

\begin{defn}
A path $P$ is defined as a sequence of GPS observations. Each observation consists of a location, i.e., (latitude, longitude), and the timestamp of submission.
\end{defn}

\begin{defn}
A trip $x(i) = (o(i),d(i),t(i),\tau(i),P(i))$ is defined as a tuple with five components where $o(i)$ denotes the origin location, $d(i)$ denotes the destination location, $t(i)$ denotes the departure time, $\tau(i)$ denotes the duration and $P(i)$ represents the corresponding path of this trip. Then, the historical trip dataset with $N$ trips can be represented as a set of trips $X = {x(i) |i = 1, 2, \cdots, N}$.
\end{defn}

In our problem, we define a trip as $(o(i),d(i),t(i),s(i))$ where $o(i)$ denotes the trip origin, $d(i)$ denotes the destination, $t(i)$ represents the departure time and $s(i)$ denotes the trajectory of this trip as a sequence of timestamped GPS readings. Our dataset consists taxi trips, each containing origin, destination, departure and arrival time and the trip trajectory which is a sequence of latitude and longitudes reported along the path. 
Real-time travel time estimation is defined as: given an origin, destination and departure time, estimate the time it takes for a vehicle to transport from the origin to the destination on a predetermined path using historical and real-time trips from the dataset. 

The path of the trip is calculated using the Dijkstra's shortest path finding algorithm \cite{yang2017origin}, and then the segments of the path are extracted. We use matrix factorization to estimate the speed of vehicles in each segment simultaneously, and then sum up all the estimated travel times for those segments.

\subsection{Matrix Factorization Model}\label{sec:mfm}
Matrix factorization is the breaking down of one matrix into a product of multiple matrices. It is well studied in mathematics and is widely used for different applications such as astronomy, data mining, bioinformatics, and signal processing.

In our problem, let $T = {\{t_{ij}\}}_{n_h * n_s}$ denote the time-segment matrix, where each cell $t_{ij}$ represents the speed of vehicles in the corresponding segment $j$ at time interval $i$, $n_h$ represents the total number of time intervals, and $n_s$ represents the total number of segments. The time and segment feature vectors can be defined as $H=[h_i]$ and $S=[s_j]$ where $h_i \in {\rm I\!R}^{n_f}$,  $s_i \in {\rm I\!R}^{n_f}$ and $n_f$ is the dimension of the feature space, for all $i=1 ... n_h$ and $j=1 ... n_s$. In this problem, $n_f$ denotes the number of hidden variables in the model that need to be estimated using historical and real-time data. Since traffic data is not completely predictable, we minimize the total loss function of $H$ and $S$ to obtain the matrix $T$.

We can define the single loss due to a single-speed as:
\begin{equation}
\label{lossFunc}
    L^2 (t,h,s)=(t- <h,s>)^2
\end{equation}

The total loss function is defined as:
\begin{equation}
    \label{TotalLoss}
    L^{emp} (T,H,S)=\frac{1}{n} \sum_{(i,j) \in P} L^2 (t_{ij},h_i,s_j)
\end{equation}

In this equation, $P$ is the index set of the known speeds and $n$ is the size of $P$. We can then calculate the low-rank approximation using the following equation:
\begin{equation}
  (H,S)= \underset{(H,S)}{arg min} \ L^{emp} (T,H,S)  
\end{equation}

\begin{equation} 
\label{eqs:ALS}
\resizebox{0.8\hsize}{!}{$
\begin{bmatrix}
t_{1}^{1} & t_{2}^{1}  & \dots  & t_{n_s}^{1} \\
t_{1}^{2} & t_{2}^{2}  & \dots  & t_{n_s}^{2} \\
\vdots & \vdots & \ddots  & \vdots \\
t_{1}^{n_t} & t_{2}^{n_t}  & \dots  & t_{n_s}^{n_t}
\end{bmatrix}
=
\begin{bmatrix}
h_{1}^{1} & h_{2}^{1}  & \dots  & h_{n_f}^{1} \\
h_{1}^{2} & h_{2}^{2}  & \dots  & h_{n_f}^{2} \\
\vdots & \vdots &  \ddots & \vdots \\
h_{1}^{n_h} & h_{2}^{n_h}  & \dots  & h_{n_f}^{n_h}
\end{bmatrix}
*
\begin{bmatrix}
s_{1}^{1} & s_{2}^{1}  & \dots  & s_{n_s}^{1} \\
s_{1}^{2} & s_{2}^{2}  & \dots  & s_{n_s}^{2} \\
\vdots & \vdots &  \ddots & \vdots \\
s_{1}^{n_f} & s_{2}^{n_f}  & \dots  & s_{n_s}^{n_f}
\end{bmatrix}
$}
\end{equation}

The matrices $H$ and $S$ resulted from factorization contain important information, the matrix $H$ represents how much each time interval is related to a given feature, and the matrix $S$ represents how much each feature is important for each segment.

To learn the two matrices $H$ and $S$, an iterative process called alternating least square (ALS) can be performed as described below.
\begin{enumerate}
  \item Initialize matrix $S$ by assigning the average speeds for that segment to the first row, and small random numbers for the remaining entries.
  \item Fix $S$, find values for $T$ by minimizing the objective function in Eq. \ref{TotalLoss}.
  \item Fix $T$ and find values for $S$ that minimize the objective function (similar to step 2).
  \item Repeat steps 2 and 3 until a certain stopping criterion is met.
\end{enumerate}

The stopping criteria is usually set by an error function. In our problem, there are $n_h \times n_f$ parameters for time interval and $n_s \times n_f$ parameters for road segments that need to be determined. If the number of parameters $n_f$ is large we may encounter the problem of overfitting the data. We use Tikhonov regularization to avoid overfitting as represented in \cite{zhou2008large}:

\begin{equation}
\label{Tikhonov_regularization}
    L_{\lambda}^{reg} (T,H,S)= L^{emp} (T,H,S) + \lambda(||U\Gamma_H ||^2+ ||U\Gamma_S ||^2)
\end{equation}

We use the weighted $ \lambda $-regularization formula described in (\ref{lambda_regularization}).
\begin{equation}
\label{lambda_regularization}
 f(H,S) = \sum_{(i,j) \in P}(t_{ij} - h_{i} s_j)^2 + \lambda (\sum_i n_{h_i} ||h_i||^2 + \sum_i n_{s_i} ||s_i||^2)   
\end{equation}

In Eq. (\ref{lambda_regularization}), $n_{h_i}$ and $n_{s_j}$ denote the number of segment-speeds for time interval $i$ and segment $j$, respectively. To determine the matrix $H$ when $S$ is given, we solve a regularized linear least-squares problem that involves the segment speeds of the time interval $i$ and the feature matrix $s_j$ for the segment speeds which are related to the time interval $i$. Similar to \cite{zhou2008large}, we can compute $h_i$ and $s_j$ as follows:

\begin{equation}
    h_i=A_i^{-1} V_i,\forall i
\end{equation}
\begin{equation}
    s_j=A_j^{-1} V_j,\forall j 
\end{equation}
\begin{equation}
    A_i=H_{P^(s)_i} H_{P^(s)_i}^T+ \lambda n_{h_i} E
\end{equation}
\begin{equation}
    V_i=H_{P^(s)_i} R^T (i, P^(s)_i)  
\end{equation}
\begin{equation}
    A_{ij}=S_{P^(t)_j} S_{P^(s)_i}^T+\lambda n_{s_i} E
\end{equation}
\begin{equation}
    V_{ij}=S_{P^(s)_i} R(P^(t)_j,j)
\end{equation}

where $P^(s)_i$ denotes the number of segments that have a segment-speed for time interval $i$  and similarly, $ P^(t)_j $ denotes the number of time intervals that had a segment-speed for segment $j$. $E$ is the $n_f \times n_f$ identity matrix, $H_(P^(s)_i)$ denotes the sub-matrix of $H$ with selected rows of $j \in P^(s)_i $, $S_(P^(t)_j)$ denotes the sub-matrix of S with selected columns of $i \in P^(t)_j$,  $R(i,P^(s)_i)$ is the columns vector where rows $j \in P^(s)_i$ of the $i$-th column of $R$ is taken and $R( P^(t)_j,j)$ is the row vector where columns $i \in P^(s)_i$ of the $j$-th row of $R$ is taken.

For this method to work, we also need to determine the number of features $n_f$. This number can be estimated through trial runs of the algorithm with different values for $H$, because for each dataset, the number of features would be different.

\subsection{Online Data}
The next challenge is updating the time-segment matrix for new time slots. The simplest way is to add new time slot traffic segments as a column to the time-segment matrix and run the matrix factorization algorithm again but it would consume high processing resources as it is shown in Eq. (5). Adding a new column to the time-segment matrix would affect the matrices $H$ and $S$, but we can assume that the old matrix $S$ could still be used for new time slots to prevent redundant heavy processing, and update $S$ after a specific number of time intervals. Using this method, only a new column would be added to the matrix $H$, also for the next time intervals, the added column values should be updated by related time slot's information.

According to Eq. \ref{eqs:Update} for estimating matrix $H$'s new row values, we can use the linear relation between matrix $S$ and matrix $T$'s rows.  

\begin{equation} 
\label{eqs:Update}
\resizebox{0.8\hsize}{!}{$
\begin{bmatrix}
t_{1}^{n_{t+1}} & t_{2}^{n_{t+1}}  & \dots  & t_{n_s}^{n_{t+1}}
\end{bmatrix}
=
\begin{bmatrix}
h_{1}^{n_{t+1}} & h_{2}^{n_{t+1}}  & \dots  & h_{n_h}^{n_{t+1}}
\end{bmatrix}
*
\begin{bmatrix}
s_{11} & s_{12}  & \dots  & s_{1 n_s} \\
s_{21} & s_{22}  & \dots  & s_{2 n_s} \\
\vdots & \vdots  & \ddots & \vdots \\
s_{n_h 1} & s_{n_h 2}  & \dots  & s_{n_h n_s}
\end{bmatrix}
$}
\end{equation}

\section{Experiments}
In this section, at first, we describe our dataset, and some preprocesses which should be done before running the experiments. After that, we introduce a method which is used for comparison with our proposed method. The performance of the two methods on the dataset would be compared and analyzed.

\subsection{Dataset}
\begin{figure}
\centering
\includegraphics[totalheight=7cm]{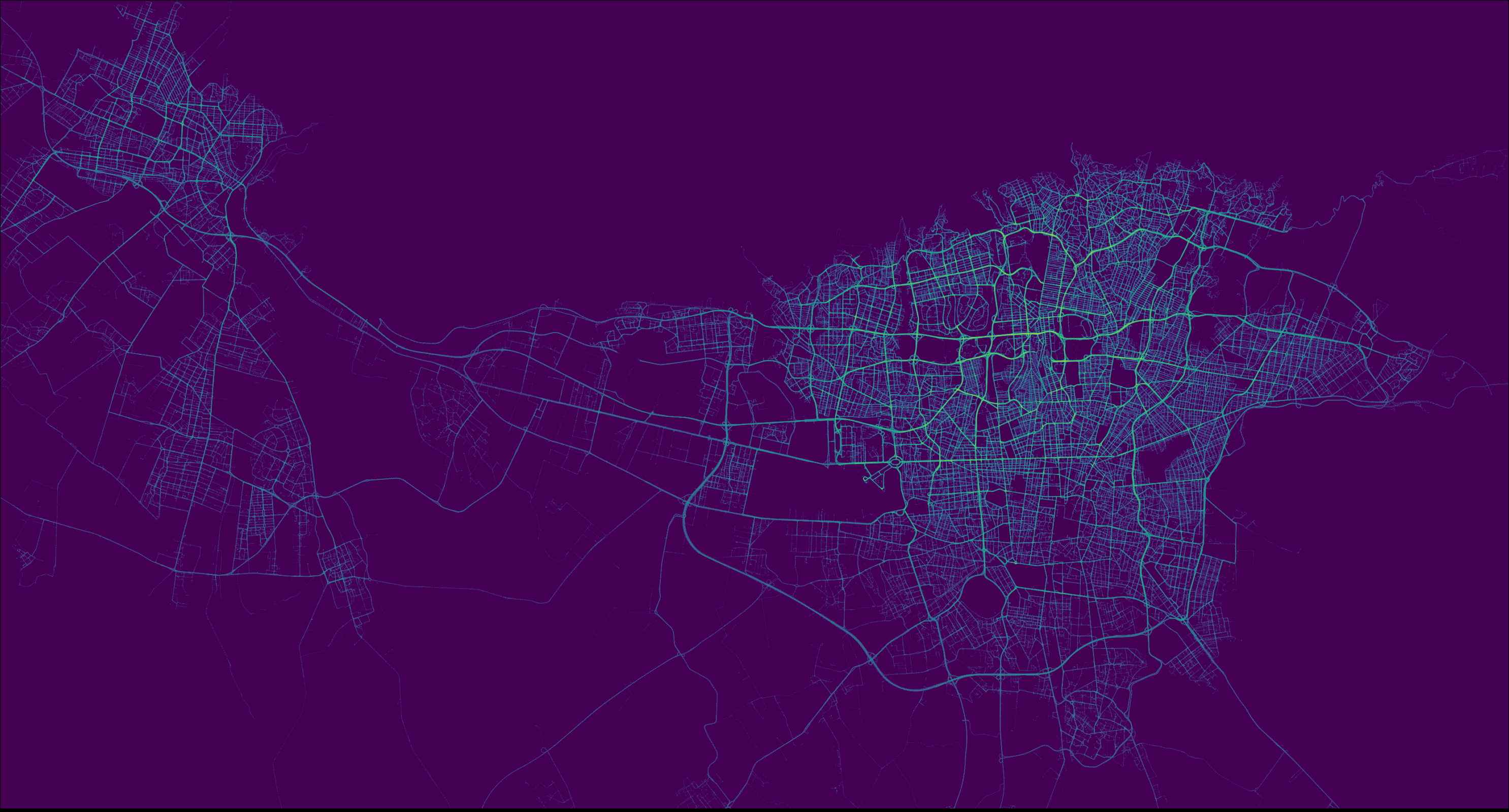}
\caption{GPS Data Distribution on One Day}
\label{fig:gps_points}
\end{figure}

To train and test our proposed model, we use passenger travel data of Tap30, an online taxi dispatch system that operates in Iran. The travel data are gathered in the city of Tehran, from October 1st to November 23rd, 2018. The data consists of the origin and destination of travels and the timestamps for each travel, and its size is about 50 Gigabytes. Furthermore, we use the GPS data of the taxis, which consist of coordinates of taxis, accuracy of the given location, and time. A sample of these GPS data points are shown in the Fig. \ref{fig:gps_points}. We use travel data of the last week in the dataset for the model validation and the rest of the data for training.

\subsection{Data Cleaning}
\begin{figure}
\centering
\includegraphics[totalheight=4cm]{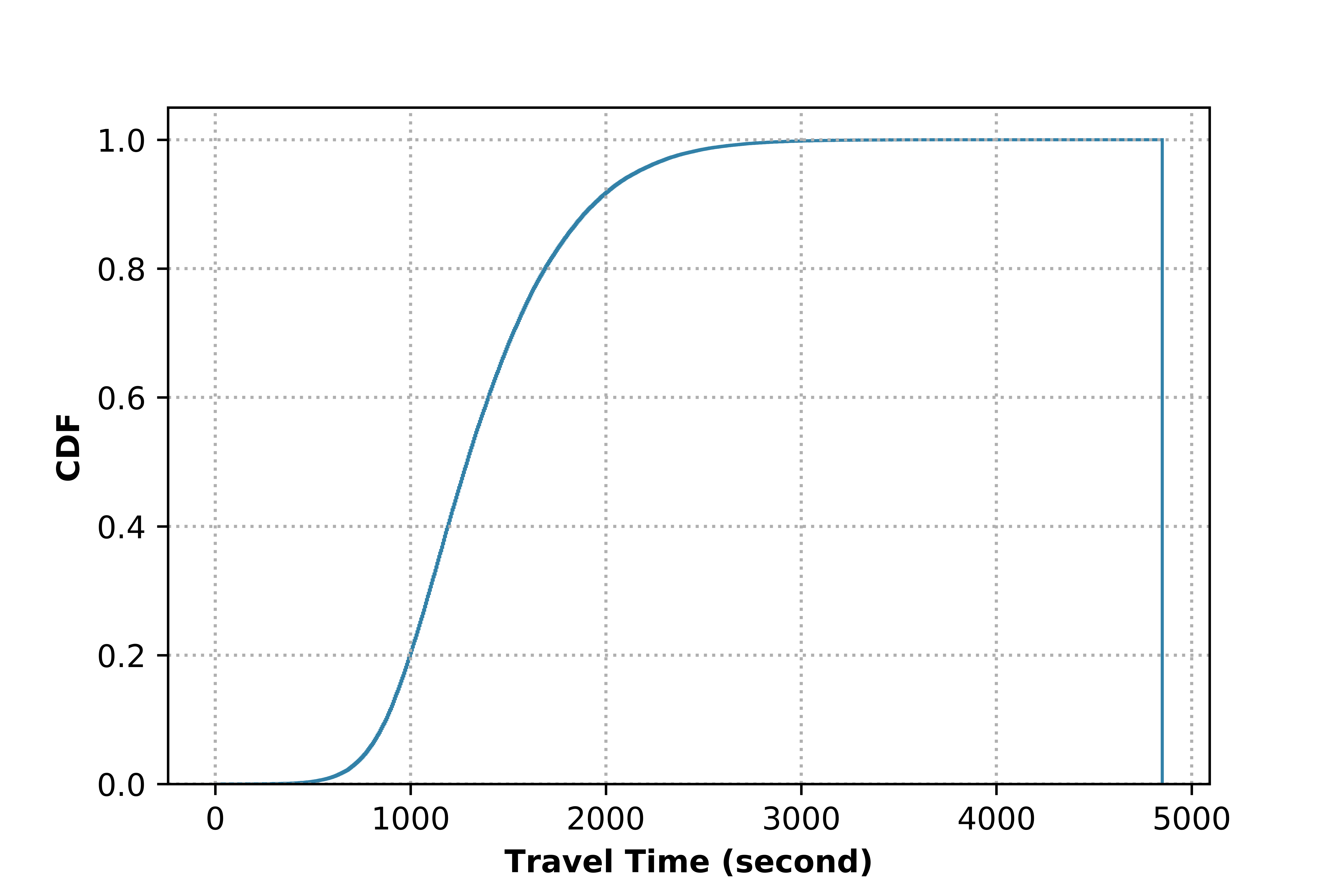}
\includegraphics[totalheight=4cm]{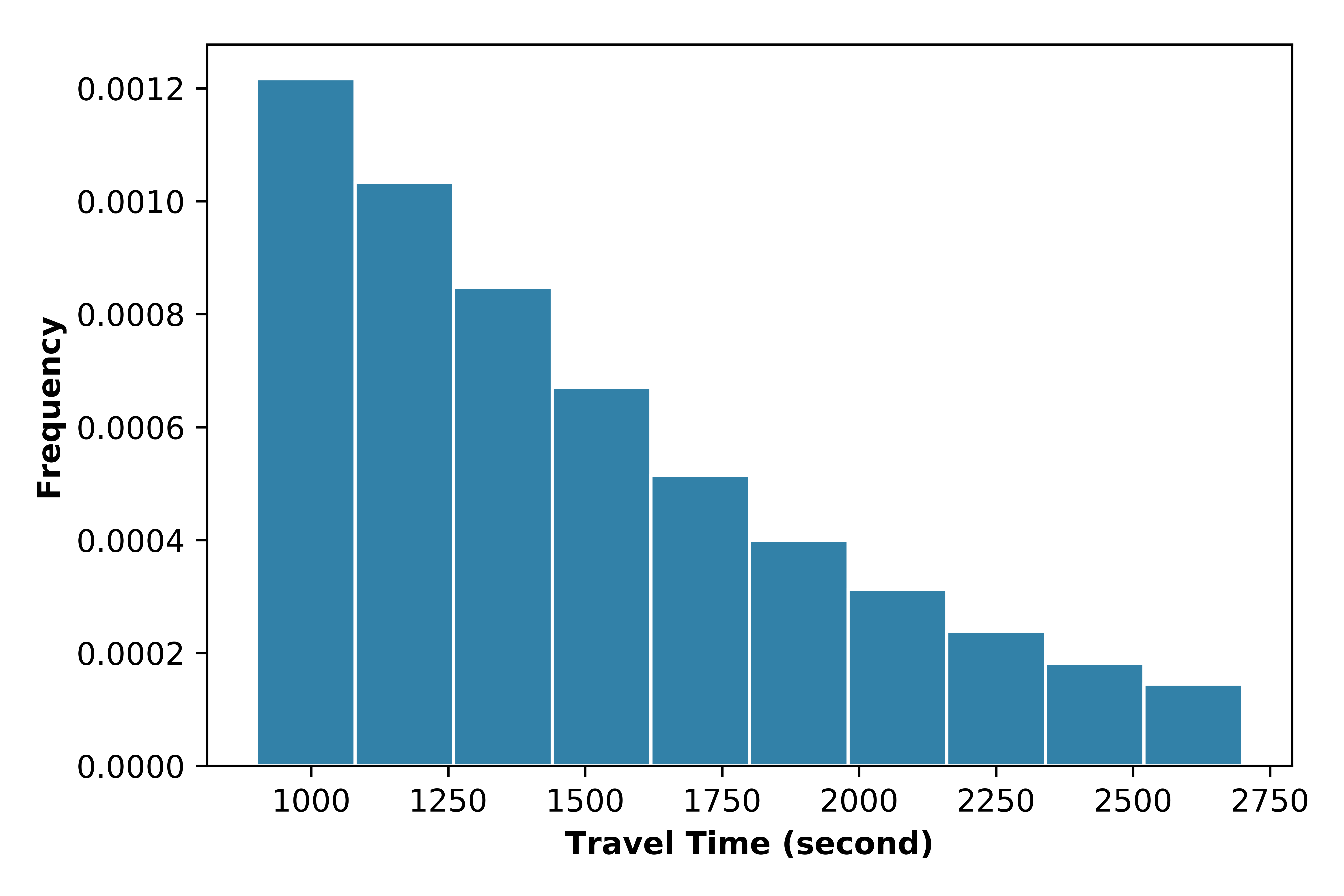}
\caption{Travel Time Statistics}
  \label{fig:travel_time_statistics}
\end{figure}

For evaluation, we only consider the trips that take from 10 to 45 minutes. Also, the trips with relevant negative comments from the passengers (e.g. "The driver selected a bad route") are ignored. About 20 percent of the trips are removed after cleaning the data.
The distribution of the travels' duration is shown in Fig. \ref{fig:travel_time_statistics}. This plot shows that about 80 percent of the travels are done in less than 30 minutes, and a few outliers had a longer duration. Most of the trips take less than 20 minutes which shows the importance of the estimation for short trips.

\subsection{Evaluation}
For evaluation and comparison of our proposed method, we implemented Hongijan's method \cite{wang2019simple} and evaluated its performance on our dataset. This method basically uses average travel time of historical trips between two points to estimate the travel time. The trips with origin and destination close to the current trip are used as historical data. Traffic variance at different times have been resolved by making use of traffic patterns. One possible advantage of this model is that as it does not use GPS data, the lack of data for some streets and tunnels does not damage its performance. 

The performance of this method on our data for different hours of the day is shown in Table 1. This table shows that our proposed method outperforms the Hongijan's method. In addition, because of the restrictions of the radius parameter value, the Hongijan's method is not able to estimate the arrival time for some trips. 

We also implemented a modified version of the Hongijan's method using a KD-Tree data structure to increase its performance, but our proposed method still has a lower response time.

\begin{table}[]
\centering
\begin{tabular}{|l|l|l|l|l|}
\hline
\textbf{Metric} & \textbf{MAPE} & \textbf{RMSE} & \textbf{MAE} & \textbf{95\%} \\ \hline
TAP30          & 16.7          & 341s          & 252s         & 41\%          \\ \hline
Hongijan       & 20.5          & 420s          & 292s         & 54\%          \\ \hline
\end{tabular}
\caption{Models' Overall Performances Comparison}
\label{factors_table}
\end{table}

Furthermore, the Mean Absolute Percentage Error (MAPE) metric is used to compare the two methods. Since a high percentage of our dataset consists of short trips, which last less than thirty minutes, MAPE, which is highly sensitivity to those trips, can measure the performance of these methods effectively. Moreover, the Mean Absolute Error (MAE) and Root Mean Squared Error (RMSE) metrics are used for further evaluation, but we believe that they are not as important as MAPE because these two metrics do not reflect errors for short trips.
\begin{equation} 
\label{MAPE@1_cost}
MAPE = \frac{100\%}{n}\sum_{t=1}^{n}\abs{\frac{e_t}{y_t}}
\end{equation}

\begin{equation} 
\label{MAE_cost}
MAE = \frac{1}{n}\sum_{j = 1}^{n}|\hat{y}_{j}-y_{j}|
\end{equation}

\begin{equation} 
\label{MSE_cost}
RMSE = \sqrt{\frac{1}{n}{\sum_{t=1}^{n}e_t^2}}
\end{equation}

\subsection{Results}
To analyze our proposed method, the results for different hours of day are compared with Hongijan's method. Results show that Hongijan's method is not able to estimate time of arrival for 524 rides out of 141121 rides used for evaluation, which is about 0.4\% of the test data. Since these rides do not have much impact on the final assessments, we remove them from the test data when comparing the two methods.

As shown in the Table \ref{factors_table}, MAPE of the two methods are compared. The values for the proposed method and Hongijan's method are 16.7 \%, 20.5 \%, respectively, which indicates that our proposed method has a lower estimation error. Based on the data shown in Table 1, the proposed method outperforms Hongijan's method in all metrics.

\begin{figure}
\centering
\includegraphics[totalheight=8cm]{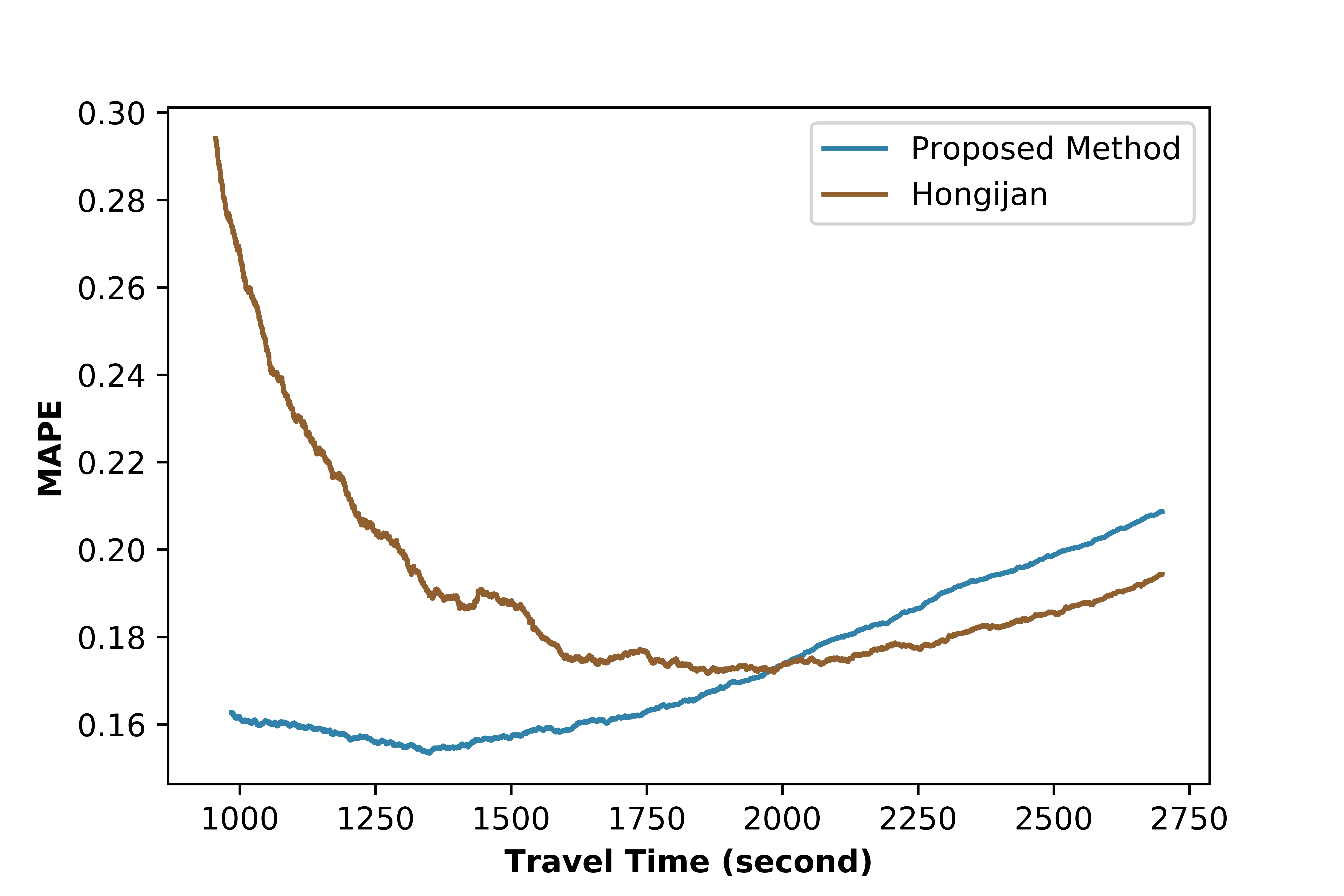}
\caption{Models Comparison w.r.t. Travel Time}
\label{fig:error_duration}
\end{figure}

Short trips have an impact on the result of the origin-destination-based methods. As shown in Fig. \ref{fig:error_duration}, Hongijan's method is more error-prone to shorter trips than longer trips which take more than 30 minutes. This method performs better for the trips that are longer than 30 minutes. Two reasons can explain this: (1) in our proposed method the route's traffic is only calculated at the beginning of the trip, but it can change as we get close to the end of the trip, (2) our method does not consider delays in some intersections, and the total error for the route is summed by the errors for each of those intersections, which adds up to larger amount for longer trips. On the other hand, none of the mentioned problems can affect the error of origin-destination-based methods because those changes and delays are also considered in similar travels. 

\begin{figure}
\centering
\includegraphics[totalheight=8cm]{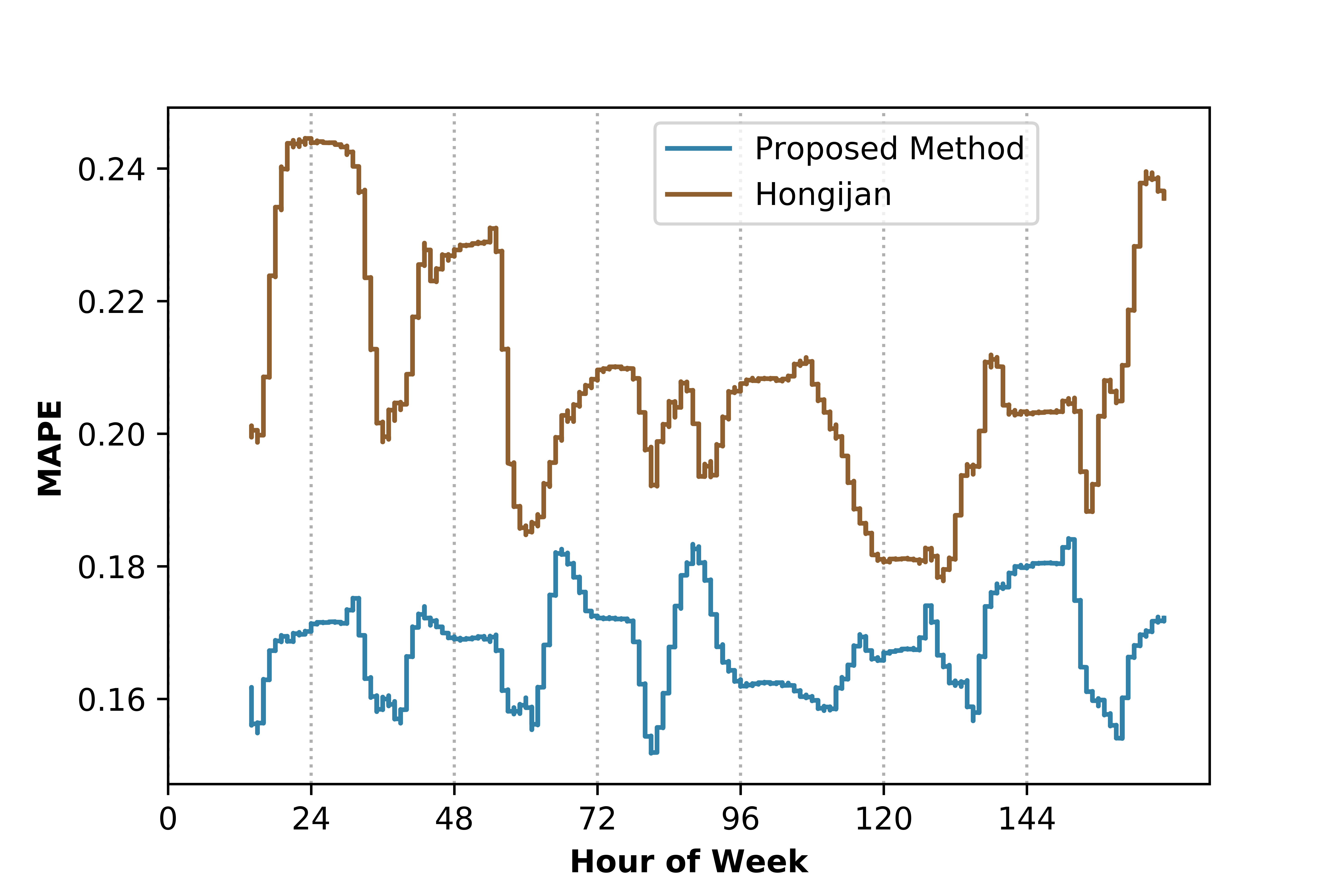}
\caption{Models Comparison w.r.t. Hour of Week}
\label{fig:error_hour_week}
\end{figure}

Fig. \ref{fig:error_hour_week} illustrates the estimation error of the models with respect to hours of the week which demonstrates the superiority of the proposed method.

\begin{figure}
\centering
\includegraphics[totalheight=8cm]{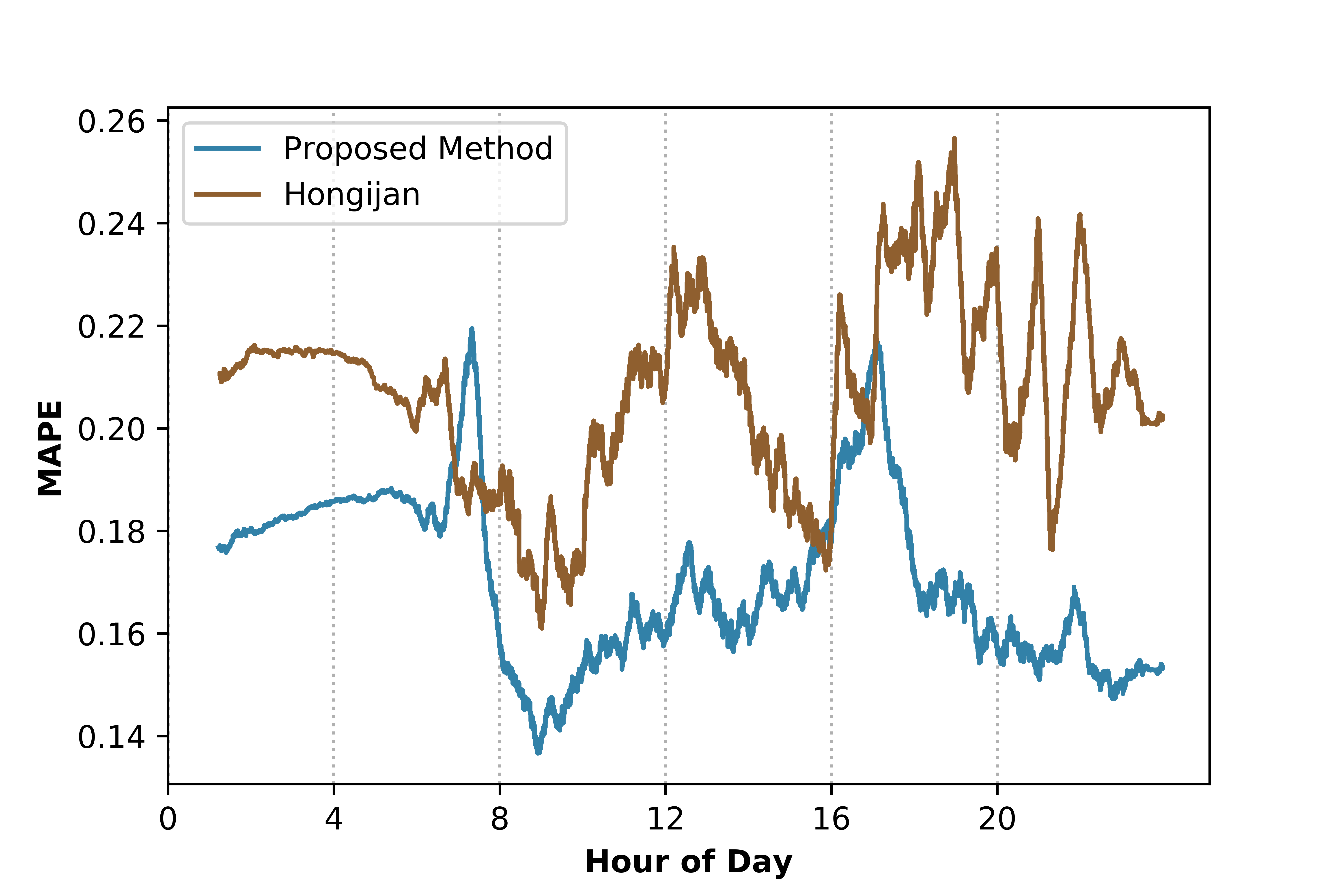}
\caption{Models Comparison w.r.t. Hour of Day}
\label{fig:error_hour_day}
\end{figure}

Estimation error comparison of the two methods for different hours of the day in Fig. \ref{fig:error_hour_day} shows that when the number of trips is high - about 8 am and 4 pm - Hongijan's method performs better or the same as the proposed method. As mentioned earlier, one reason for this difference is that our proposed method calculates the traffic at the beginning of the trip, and at these hours traffic congestion is highly variable in Tehran. But for other hours of the day, our method can achieve higher accuracy.

\begin{figure}
\centering
\includegraphics[totalheight=8cm]{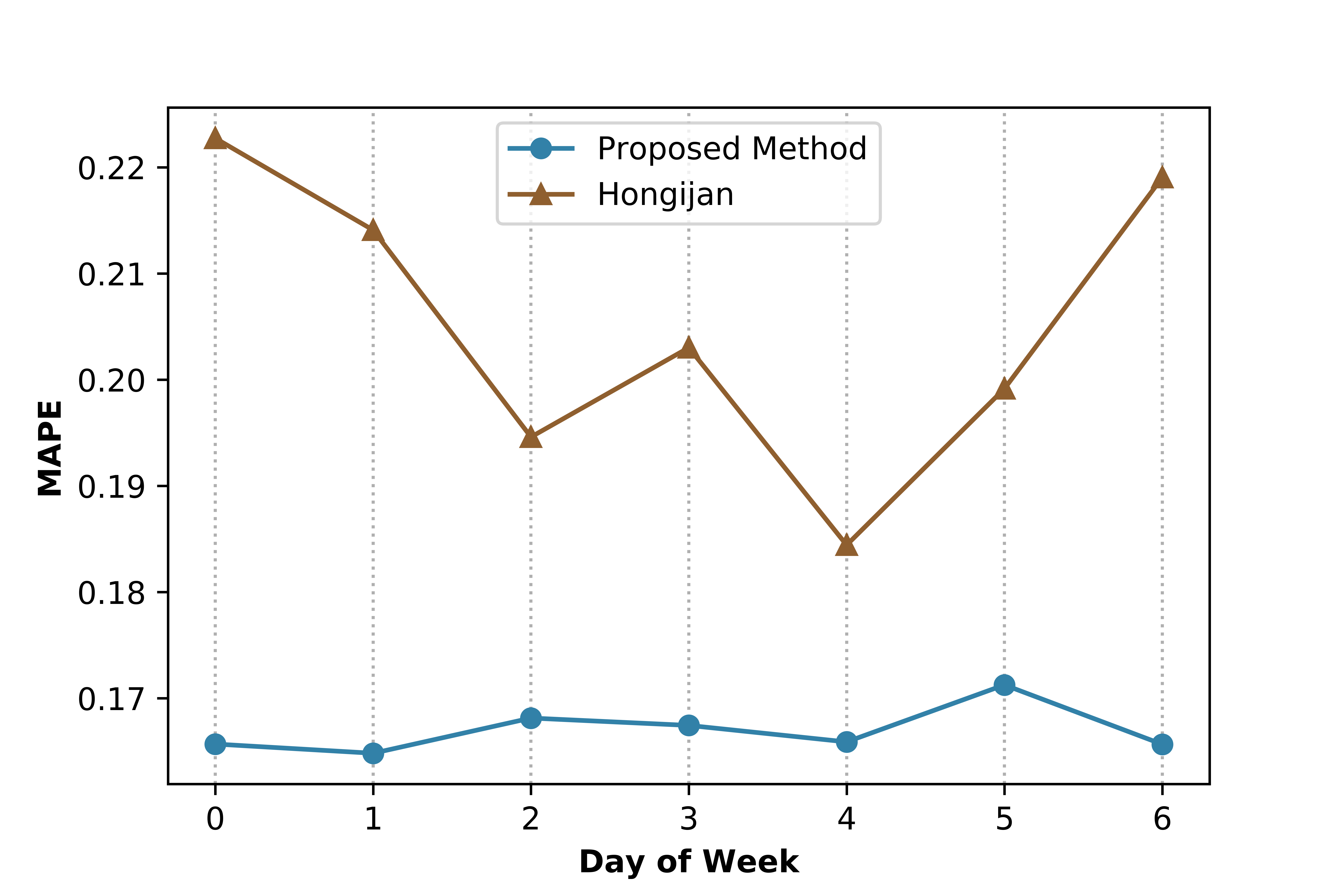}
\caption{Models Comparison w.r.t. Day of Week}
\label{fig:error_day}
\end{figure}

Moreover, the models' estimation error aggregation for days of a week is demonstrated in the Fig. \ref{fig:error_day} which shows the robustness of our method. In fact, Hongijan's method's error fluctuates much more on different days.

As the experiments show, our proposed method's estimation accuracy is better than Hongijan's method for short trips which are a significant proportion of the trips. The fact which makes the Hongijan's method more desirable is its independence of GPS data. If the GPS data were not available, Hongijan's method would be preferred. But, providing with those data, our proposed method has a better accuracy, less response time, and more robustness.

\label{S:5}
\section{Conclusion}
In this work, we proposed a real-time and scalable model for estimating time of arrival using online massive dense GPS trajectories. The number of road segment in a city is so large that sparsity and noise are the main problems for estimating the speed of each road segment, independently. Thus, we used a matrix factorization algorithm called ALS-WR to reduce the dimensionality of the problem.

We used the duration of real rides of an online taxi platform as the ground truth for evaluating our model and calculated some error factors like MAPE over it. We also implemented Hongijan’s method to compare our result with it. As shown in the experiment section, our method outperforms Hongijan’s, in many ways including accuracy, response time, scalability, robustness to amount of available data, and robustness to unpredictable events since we use online stream of data. But there are also some advantages for Hongijan’s method over ours, mainly accuracy of travel time estimation for long trips. 
\label{S:6}\label{S:5}





\bibliographystyle{model1-num-names}
\bibliography{sample}







\end{document}